\documentclass[10pt,twocolumn,letterpaper]{article}

\usepackage{iccv}
\usepackage{times}
\usepackage{epsfig}
\usepackage{graphicx}
\usepackage{amsmath}
\usepackage{amssymb}
\usepackage{algorithm}
\usepackage{algorithmic}
\usepackage{bbm}
\usepackage{amsfonts}
\usepackage{mathrsfs}
\usepackage{multirow}
\usepackage{amsmath,amssymb,amsfonts}
\usepackage{algorithmic}
\usepackage{graphicx}
\usepackage{textcomp}
\usepackage{booktabs}
\usepackage{multicol}
\usepackage{booktabs}
\usepackage{lipsum}
\usepackage{makecell}
\usepackage{caption}
\usepackage{multirow}
\usepackage{placeins}

\usepackage[pagebackref=true,breaklinks=true,letterpaper=true,colorlinks,bookmarks=false]{hyperref}

\iccvfinalcopy 

\def\iccvPaperID{406} 

\begin{document}

\title{ESICA: A Scalable Framework for Text-Guided 3D Medical Image Segmentation}

\author{Yu Xin{$^{1}$},~ Gorkem Can Ates{$^{1}$},~ Jun Ma{$^{2}$},~ Sumin Kim{$^{2}$},~ Ying Zhang{$^{1}$},\\
Kaleb E Smith{$^{3}$},~ Kuang Gong{$^{1}$},~ Wei Shao{$^{1*}$}\\\vspace{-8pt}{\small~}\\
{$^{1}$}University of Florida,~ {$^{2}$}University of Toronto,~ {$^{3}$}NVIDIA Corporation\\
{\small{{$^{*}$}Corresponding to: \tt{weishao@ufl.edu}}}
}

\maketitle
\footnotetext[1]{This work has been submitted to the IEEE for possible publication. Copyright may be transferred without notice, after which this version may no longer be accessible.}
\begin{abstract}
Text-guided 3D medical image segmentation offers a flexible alternative to class-based and spatial prompt-based models by allowing users to specify regions of interest directly in natural language. This paradigm avoids reliance on predefined label sets, reduces ambiguous outputs, and aligns more naturally with clinical workflows. However, existing text-guided frameworks are often computationally expensive, exhibit weak text–volume feature alignment, and fail to capture fine anatomical details. We propose ESICA, a lightweight and scalable framework that addresses these challenges through three innovations: (1) a similarity-matrix-based mask prediction formulation that enhances semantic alignment, (2) an efficient decomposed decoder with adapter modules for accurate volumetric decoding, and (3) a two-pass refinement strategy that sharpens boundaries and resolves uncertain regions. To improve training stability and generalization, ESICA adopts a two-stage scheme consisting of positive-only pretraining followed by balanced fine-tuning. On the CVPR-BiomedSegFM benchmark spanning five imaging modalities (CT, MRI, PET, ultrasound, and microscopy), ESICA achieves state-of-the-art segmentation accuracy, while the compact ESICA4-Lite variant attains similar segmentation performance with substantially fewer parameters, yielding a superior efficiency–accuracy trade-off. Our framework advances text-guided segmentation toward efficient, scalable, and clinically deployable systems. Code will be made publicly available at \url{https://github.com/mirthAI/ESICA}.
\end{abstract}

\section{Introduction}
\label{sec:introduction}
3D medical image segmentation plays a central role in computer-assisted diagnosis and treatment by delineating organ boundaries and abnormalities. It supports critical clinical tasks such as surgical planning, radiation therapy, and disease monitoring with improved precision and consistency. However, conventional segmentation methods rely heavily on voxel-wise annotations and predefined class labels, which limit scalability and generalizability. Creating voxel-level annotations requires expert radiologists to outline structures slice by slice, rendering the process time-consuming, costly, and impractical at scale. Moreover, dependence on fixed label sets restricts the ability of models to generalize to previously unseen anatomies, pathologies, or imaging modalities without substantial retraining. These limitations hinder the real-world applicability of fully supervised segmentation pipelines.

To alleviate the limitations of fully supervised segmentation, spatial prompt-based approaches have emerged as a partial step toward more flexible segmentation. These methods enable models to generate segmentations conditioned on user-provided spatial cues, such as points and bounding boxes, thereby reducing reliance on dense voxel-wise annotations.
The Segment Anything Model (SAM) \cite{SAM-ICCV23} demonstrated that large-scale pretraining enables effective interactive segmentation using spatial prompts. Subsequent medical adaptations  \cite{SAM-Med3D, MedSAM, MedSAM2} extended this framework to 3D volumetric data and incorporated interaction mechanisms better aligned with clinical imaging scenarios. Despite these advances, spatial prompt–based methods remain constrained by their reliance on precise manual inputs. Providing accurate points or bounding boxes is time-consuming and requires domain expertise, limiting practicality in routine clinical workflows. Moreover, spatial cues can be ambiguous or insufficient when anatomical boundaries are unclear, motivating the need for more expressive and flexible forms of interaction.

Text-guided segmentation has emerged as a promising alternative to spatial prompt-based methods, offering greater flexibility and scalability. By allowing clinicians to describe regions of interest in natural language, these approaches can adapt to unseen structures, rare conditions, and task-specific variations without requiring new annotations. Compared to spatial prompts that rely on manual inputs such as bounding boxes or clicks, text-based guidance reduces annotation effort and mitigates ambiguity when anatomical boundaries are unclear. For instance, a radiologist may simply specify “cystic lesion in the right kidney,” making the interaction more intuitive and better aligned with clinical workflows. In addition to improved usability, text supervision eliminates the need to harmonize class definitions across datasets, enabling models to scale more effectively across heterogeneous sources and complex clinical tasks.

Recent methods such as CAT \cite{CAT} and SAT \cite{SAT} have demonstrated the feasibility of natural language–guided 3D segmentation across organs and imaging modalities. In parallel, Text3DSAM \cite{xin2025text3dsam} adapts the SAM-style promptable segmentation paradigm to text-guided 3D segmentation.
Despite these advances, existing approaches remain computationally intensive and often struggle with fine-grained alignment between textual and volumetric features. As a result, their coarse predictions may fail to capture small or irregular anatomical structures and are often impractical for deployment in resource-constrained clinical environments. These limitations underscore the need for frameworks that can simultaneously enhance semantic alignment, boundary precision, and computational efficiency, while supporting both lightweight and full-capacity variants for diverse deployment scenarios.

In this paper, we introduce a novel framework for text-guided 3D medical image segmentation that addresses key limitations of existing approaches. Our model is offered in two configurations: a lightweight variant (130M parameters) suitable for resource-constrained deployment, and a standard variant (252M parameters) optimized for maximum accuracy. To improve semantic alignment, we reformulate mask prediction as a similarity matrix between visual and textual embeddings, enabling the model to generate masks that more faithfully reflect prompt's intent. This formulation enforces strong image–text consistency using only Dice–Focal loss, reducing reliance on auxiliary loss terms as in CAT \cite{CAT} and SAT \cite{SAT} and resulting in shorter training times. For efficient volumetric decoding, we design a new decoder based on EffiDec3D \cite{rahman2025effidec3d}, incorporating the decomposed block from DCFormer \cite{ates2025dcformer}. To further enhance segmentation quality, we introduce a two-pass refinement strategy that iteratively sharpens boundaries and resolving uncertain regions. Training stability and generalization are achieved through a two-stage curriculum, starting with positive-only pretraining followed by balanced fine-tuning.

Our main contributions are as follows:
\begin{itemize}
  \item We propose a versatile framework for text-guided 3D image segmentation that integrates efficient architectural design with resource-aware volumetric decoding.
  \item We introduce a similarity matrix–based mask prediction formulation and a two-pass refinement strategy, which together enhance semantic alignment, improve boundary precision, and reduce computational overhead.
  \item We develop a progressive training scheme that first pretrains with positive-only supervision for stability, followed by balanced positive–negative fine-tuning for improved generalization.
  \item We conduct extensive experiments across CT, MRI, PET, ultrasound, and microscopy, demonstrating consistent improvements in both accuracy and efficiency over prior text-guided baselines.
\end{itemize}

\section{Related Work}
\label{section2}
\subsection{Fully Supervised Medical Image Segmentation}

Traditional medical image segmentation methods are typically fully supervised and require dense, labor-intensive annotations. U-Net \cite{ronneberger2015u} and its variants remain among the most widely used architectures, demonstrating strong performance across modalities such as CT, MRI, and ultrasound. Subsequent work extended U-Net to 3D \cite{cciccek20163d} and incorporated attention mechanisms \cite{oktay2018attention, jha2019resunet++, wang2022uctransnet} and vision transformers \cite{chen2021transunet, cao2022swin, he2023swinunetr} to improve modeling capacity and generalization. While effective, these models rely heavily on large-scale, manually annotated datasets, which are expensive and time-consuming to obtain. In addition, their limited ability to generalize to new tasks or domains often necessitates retraining or fine-tuning, which reduces scalability in real-world clinical workflows.

\subsection{Spatial Prompt-Based Segmentation}
In recent years, prompt-based segmentation frameworks have emerged as a promising alternative for reducing annotation costs and increasing flexibility. The Segment Anything Model (SAM) \cite{SAM-ICCV23} introduced large-scale pretraining for interactive segmentation using spatial prompts such as points, bounding boxes, and masks. Several medical adaptations have extended SAM to 3D volumetric imaging. For example, MedSAM \cite{MedSAM} and MedSAM2 \cite{MedSAM2} demonstrated improved cross-task generalization in medical contexts, while Efficient MedSAM \cite{ma2024efficient} optimized the framework for more efficient training and inference pipelines. RepViT-MedSAM \cite{ali2024repvit} introduced a lightweight re-parameterized vision transformer backbone, enabling faster inference and reduced memory usage without sacrificing accuracy. Similarly, DSAM \cite{tan2024dsam} proposed a streamlined 3D adaptation that lowers computational cost and latency while maintaining strong performance across volumetric modalities. FastSAM3D \cite{FastSAM3D} further accelerated inference through layer-wise progressive distillation and sparse 3D flash attention. Other approaches, such as SegVol \cite{segvol}, have explored volumetric interactive segmentation through spatial prompts and multimodal fusion. Despite these advances, spatial prompt--based methods still require clinicians to provide accurate manual inputs, such as bounding boxes or clicks, which can be impractical in routine workflows and ambiguous when anatomical boundaries are unclear.

\subsection{Text-Guided Segmentation}

Recent studies have explored text as a natural and expressive form of supervision for medical image segmentation. CLIPSeg \cite{luddecke2022image} first demonstrated text-to-mask prediction in natural images using CLIP embeddings and has since served as a foundation for medical adaptations. CAT \cite{CAT} coordinates anatomical and textual prompts to support multi-organ and tumor segmentation, while SAT \cite{SAT} extends this paradigm toward universal text-guided segmentation across imaging modalities. Text3DSAM \cite{xin2025text3dsam} adapts the SAM-style promptable segmentation framework to 3D medical volumes with language prompts, providing an efficient reference for text-to-mask prediction in volumetric data. BioMedParse \cite{BioMedParse} introduced a large-scale biomedical foundation model capable of joint segmentation, detection, and recognition across nine imaging modalities. Despite these advances, existing frameworks remain computationally demanding and often exhibit weak text–volume alignment, limiting their scalability and precision in clinical settings.

\section{Method}
\label{section3}

\begin{figure*}[!hbt]
  \centering
  \includegraphics[width=\textwidth]{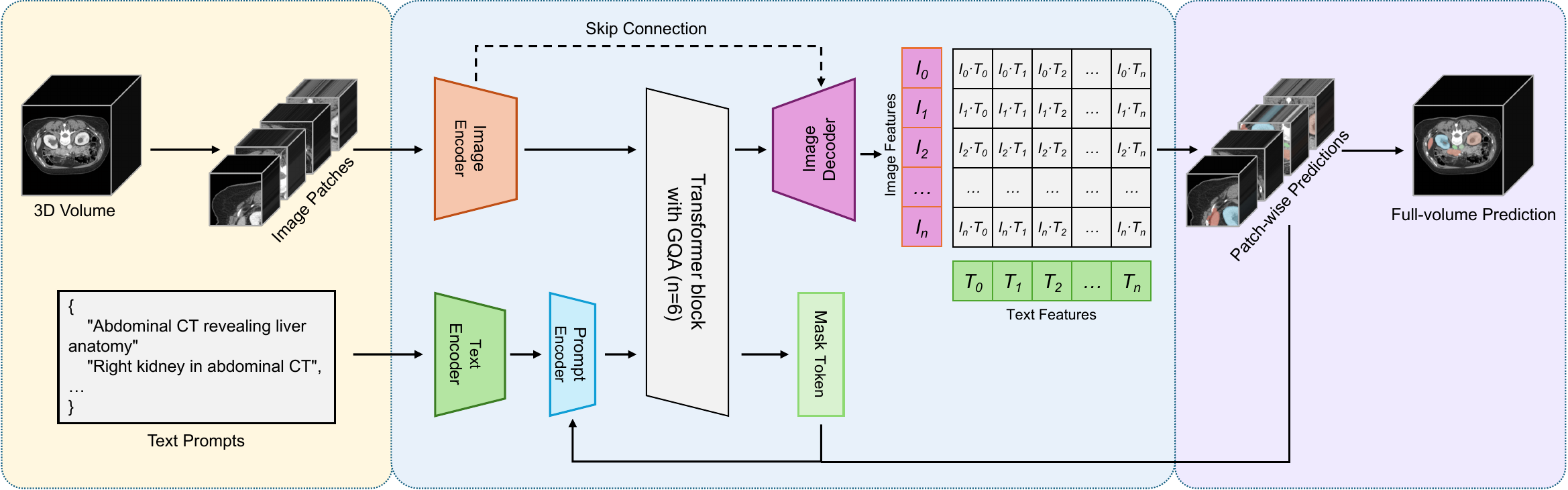}
  \caption{Overview of the proposed ESICA framework. A 3D volume is partitioned into patches and embedded by an image encoder, while free-text prompts are processed by a parallel text encoder. The core transformer module fuses these multimodal features and enables iterative refinement via a prompt encoder and mask token. Patch-wise masks are reconstructed by the decoder and aligned with text features through a similarity matrix. Final full-volume predictions are obtained by aggregating the patch-level outputs.}
  \label{fig:pipeline}
\end{figure*}

\subsection{Proposed Method: ESICA}
We introduce \textbf{ESICA}: \textbf{E}fficient \textbf{S}egmentation with \textbf{I}terative refinement and \textbf{C}ross-modal \textbf{A}lignment, a novel framework for text-guided 3D medical image segmentation. As shown in Figure \ref{fig:pipeline}, the 3D input volume is partitioned into a sequence of patches, which are processed by five key components: an image encoder, a text encoder, a prompt encoder, a transformer module, and an efficient decoder. The prompt encoder enables iterative refinement by fusing text embeddings with the mask token and features from the previous prediction. The transformer module performs cross-modal reasoning between image, text, and prompt tokens via multi-head cross-attention. The decoder outputs patch-wise segmentation masks, which are reassembled into a full-volume mask. The final output is produced by aligning upscaled image features with text features through a similarity matrix.

\subsubsection{Image Encoder}
We propose \textbf{DCFormerV2} as our image encoder, extending the original DCFormer~\cite{ates2025dcformer}, which was designed for efficient 3D vision–language modeling. DCFormer factorizes standard 3D convolutions into three depthwise branches along the height, width, and depth axes, significantly reducing FLOPs and parameter counts while maintaining global context modeling. Our key modification lies in the decomposed convolution block. In addition to the original three parallel 1D convolutions ($k \times 1 \times 1$, $1 \times k \times 1$, and $1 \times 1 \times k$), we introduce a fourth branch with a standard $3 \times 3 \times 3$ convolution. This hybrid design enables the model to simultaneously capture fine-grained local details via the compact kernel and broader contextual cues via the decoupled convolutions. Despite this addition, the increase in model size is negligible; for example, the parameter count rises only slightly from 37.11M in DCFormer to 37.13M in DCFormerV2 under the same configuration, while the computational cost of the image encoder increases from 21.94 G to 53.04 G FLOPs. Considering the clear improvements in downstream segmentation accuracy, this modest increase in model size and compute budget is well justified by the resulting performance gains.

The operations in a DCFormerV2 block can be formulated as:
\begin{align}
  &X^*_{\text{h}} = \text{DWConv}^{k_h \times 1 \times 1}_{C \rightarrow C} (X), \\
  &X^*_{\text{w}} = \text{DWConv}^{1 \times k_w \times 1}_{C \rightarrow C} (X), \\
  &X^*_{\text{d}} = \text{DWConv}^{1 \times 1 \times k_d}_{C \rightarrow C} (X), \\
  &X^*_{\text{local}} = \text{DWConv}^{3 \times 3 \times 3}_{C \rightarrow C} (X), \\
  &X^* = \text{Activation}(X^*_{\text{h}} + X^*_{\text{w}} + X^*_{\text{d}} + X^*_{\text{local}})
\end{align}

We also modify the macro-architecture to better accommodate patch-based inputs. To preserve spatial resolution, we revise the stem stage: instead of four consecutive blocks with a downsampling factor of 4, we use a single DCFormerV2 block without any downsampling. Finally, to improve long-range feature modeling, we append a standard 3D Transformer layer after the final stage of the DCFormerV2 backbone.

\subsubsection{Text Encoder}
We employ a pretrained BERT \cite{devlin2019bert} model to convert textual prompts into high-dimensional feature vectors. The input text is first processed by the corresponding pretrained tokenizer, which generates input IDs and an attention mask. These are then fed into the BERT model to extract contextualized embeddings. Following standard practice, we utilize the feature vector corresponding to the special \texttt{[CLS]} token as the final text feature representation. This token is designed to aggregate the global semantic information of the entire input sequence, providing a compact yet comprehensive representation. By using only the \texttt{[CLS]} embedding, we effectively reduce the length of the text features, which enhances computational efficiency for subsequent cross-modal fusion.

\subsubsection{Prompt Encoder}
Our prompt encoder generates the sparse and dense prompt embeddings used by the subsequent transformer blocks. At the first iteration, the sparse embedding is initialized directly from the text encoder output, while the dense embedding is represented by a set of learnable parameters, since no mask prediction is yet available.

In later iterations, the sparse embedding is formed by concatenating the text embedding with the mask token produced in the previous iteration, allowing the model to explicitly condition on prior predictions. The dense embedding is derived from the segmentation mask predicted in the preceding iteration. This mask is passed through a lightweight feature extractor composed of decomposed convolution layers, which progressively downsample the mask and project it into the shared embedding space.

This design enables the prompt encoder to fuse semantic guidance from text with spatial feedback from intermediate segmentation results. By updating prompt representations iteratively, the model progressively refines its predictions, leading to more accurate and stable segmentation outputs.

\subsubsection{Transformer Module with Grouped Query Attention}

The transformer module is designed to effectively fuse image features with prompt embeddings. The inputs are constructed as follows: the keys (K) and values (V) are obtained by element-wise addition of the image features and dense prompt embeddings, while the queries (Q) are formed by concatenating learnable mask token embeddings with the sparse prompt embeddings.

The module consists of a stack of two-way attention blocks, inspired by the SAM architecture~\cite{SAM-ICCV23}. Each block enables bidirectional information exchanges through three steps: (1) self-attention over the query tokens, (2) cross-attention where queries attend to keys and values (Q $\rightarrow$ KV), and (3) reverse cross-attention where keys attend to queries (KV $\rightarrow$ Q).

To improve computational efficiency and incorporate positional information, we introduce two modifications. First, we adopt Grouped-Query Attention (GQA)~\cite{ainslie2023gqa}, in which the $H_q$ query heads are partitioned into $G$ groups ($1 \leq G < H_q$), with each group sharing a single key and value head. This design reduces memory and computation while preserving representational capacity, which is particularly beneficial given the relatively large number of query tokens. Second, we apply Rotary Position Embeddings (RoPE)~\cite{su2024roformer} to queries and keys prior to attention. For a vector $x \in \{q, k\}$ at position $m$, RoPE rotates each feature pair as:

\begin{align}
  \begin{pmatrix} x'_{2i} \\ x'_{2i+1}
  \end{pmatrix} &=
  \begin{pmatrix} \cos(m\theta_i) & -\sin(m\theta_i) \\ \sin(m\theta_i) & \cos(m\theta_i)
  \end{pmatrix}
  \begin{pmatrix} x_{2i} \\ x_{2i+1}
  \end{pmatrix}, \nonumber\\
  \text{where } \theta_i &= 10000^{-2i/d}.
\end{align}

Here, $(x_{2i}, x_{2i+1})$ denotes a feature pair, and $d$ denotes the feature dimension.

After passing through the stack of two-way attention blocks, the updated query embeddings are fed into a final cross-attention layer (Q $\rightarrow$ KV) and then forwarded to the segmentation decoder.

\subsubsection{Image Decoder}

Our image decoder follows a U-Net-like architecture that progressively upsamples feature embeddings from the transformer module, restoring high-level semantic representations to fine-grained spatial detail. Multi-scale skip connections from the image encoder are integrated throughout the decoding path to preserve and exploit low-level spatial information.

A central component of the decoder is an efficient adapter module, inspired by EffiDec3D~\cite{rahman2025effidec3d}, which processes each skip connection prior to fusion. Unlike the original design, our adapters use DCFormerV2 blocks~\cite{ates2025dcformer}, improving computational efficiency while refining encoder features. These adapters align multi-scale feature channels to a unified hidden dimension and enhance contextual representations. To further strengthen cross-scale modeling, larger convolutional kernels are assigned to adapters operating on higher-resolution feature maps.

At each stage of the upsampling path, deeper-layer feature maps are upsampled via trilinear interpolation, refined using a $3 \times 3 \times 3$ convolution, and concatenated with the corresponding skip-connection features. The fused features are then processed by a convolutional layer to ensure effective feature propagation. This process is repeated until a high-resolution feature map, referred to as the \emph{upscaled image embedding}, is obtained.

Rather than directly producing a segmentation mask, the decoder outputs a dense image feature embedding. The final segmentation is computed via a similarity operation between the upscaled image embedding and the text embedding.
Specifically, the text embedding is projected by a multilayer perceptron (MLP) to match the channel dimension of the image features. The dot product between the aligned text vector and each spatial image feature vector yields the predicted mask. Given an upscaled image feature map $\mathbf{F}_{img} \in \mathbb{R}^{C \times H \times W \times D}$ and a text feature vector $\mathbf{F}_{text}$, the mask $\mathbf{M}$ is calculated as:

\begin{align}
  &\mathbf{F}'_{text} = \text{MLP}(\mathbf{F}_{text}), \\
  &\mathbf{M}_{h,w,d} = \langle \mathbf{F}'_{text}, \mathbf{F}_{img, h,w,d} \rangle,
\end{align}

where $\mathbf{F}'_{text} \in \mathbb{R}^{C}$ is the dimension-aligned text feature, $\mathbf{F}_{img, h,w,d} \in \mathbb{R}^{C}$ is the image feature vector at spatial coordinate $(h,w,d)$, and $\langle \cdot, \cdot \rangle$ denotes the dot product.

\subsection{Dataset}

We evaluate our method on the CVPR-BiomedSegFM dataset\footnote{https://huggingface.co/datasets/junma/CVPR-BiomedSegFM}, introduced in the CVPR 2025 paper \emph{Foundation Models for Text-guided 3D Biomedical Image Segmentation}. This benchmark aggregates 35,792 preprocessed 3D volumes collected from 68 sub-datasets spanning five imaging modalities: Computed Tomography (CT), Magnetic Resonance Imaging (MRI), 3D Ultrasound (3DUS), Positron Emission Tomography (PET), and Microscopy. For each sub-dataset, every semantic class is paired with one or more textual prompts, enabling systematic evaluation of text-guided segmentation performance.

The diversity of CVPR-BiomedSegFM facilitates comprehensive cross-modal generalization, covering a wide range of imaging domains from radiological modalities to microscopy. In all experiments, we use the official training set provided by the benchmark. Table~\ref{tab:dataset} summarizes the dataset statistics, including training and validation splits across all modalities.

\begin{table}[!ht]
  \centering
  \small
  \caption{Dataset statistics for CVPR-BiomedSegFM across five imaging modalities.}
  \label{tab:dataset}
  \begin{tabular}{lcc}
    \toprule
    \textbf{Modality} & \textbf{Training Cases} & \textbf{Validation Cases} \\
    \midrule
    CT          & 24,786 & 906 \\
    MRI         & 8,975  & 977 \\
    3DUS        & 1,122  & 110 \\
    PET         & 623    & 69 \\
    Microscopy  & 286    & 83 \\
    \midrule
    \textbf{Total} & 35,792 & 2,145 \\
    \bottomrule
  \end{tabular}
\end{table}

\subsection{Evaluation Metrics}

We evaluate model performance on two tasks: semantic and instance segmentation. For structures annotated at the class level, the task is treated as \emph{semantic segmentation}. For structures annotated at the instance level (e.g., lesions or tumors), the task is treated as \emph{instance segmentation}.

\subsubsection{Semantic Segmentation}
For class-level targets, we report:
\begin{itemize}
  \item \textbf{Dice Similarity Coefficient (DSC)}: Quantifies volumetric overlap between predictions and ground truth, based on voxel-wise true positives, false positives, and false negatives.
  \item \textbf{Normalized Surface Dice (NSD)}: Measures the fraction of boundary points correctly predicted within a tolerance $\tau$, set to 2 mm. Predicted surface points are counted as correct if they lie within $\tau$ of the ground-truth surface.
\end{itemize}

\subsubsection{Instance Segmentation}
For instance-level targets, each connected component is treated as a distinct object. When the model produces a binary mask, we perform 3D connected-component analysis to assign instance labels, and evaluation is based on matching predicted to ground-truth instances.
\begin{itemize}
  \item \textbf{F1 Score}: A prediction is a true positive if it matches a ground-truth instance with an Intersection-over-Union (IoU) $\geq 0.5$. Matching is solved with the Hungarian algorithm to enforce one-to-one correspondence. Unmatched predictions are false positives, and unmatched ground-truth instances are false negatives.
  \item \textbf{DSC for True Positives (DSC-TP)}: Evaluates segmentation quality of correctly detected instances. Dice scores are computed for each matched pair, and DSC-TP is the average over all matched pairs.
\end{itemize}

\subsection{Implementation Details}

We adopt modality-specific preprocessing strategies using the MONAI framework~\cite{cardoso2022monai}. For CT, MRI, PET, and 3DUS data, images and labels are resampled to an isotropic voxel spacing of $1.5 \times 1.5 \times 1.5$ mm, followed by foreground cropping and intensity normalization to $[0, 1]$. For microscopy data, we skip resampling and directly apply cropping and normalization. During training, we employ a patch-based sampling strategy: inputs are converted to channel-first format, padded to the target size, and cropped into fixed-size ROIs of $96 \times 96 \times 96$. To ensure balanced sampling, we use a foreground–background aware cropping scheme, where a proportion of patches are drawn around anatomical structures and the remainder from background regions.

For the text encoder, we experiment with both ClinicalBERT~\cite{wang2023optimized,liu2025generalist} and TinyClinicalBERT~\cite{rohanian2023lightweight}, providing alternatives depending on computational budget and application requirements. The transformer module contains 6 layers, with grouped-query attention configured using 12 query heads and 4 key–value heads for efficient cross-modal interaction. Iterative refinement is performed with two forward passes during both training and inference: the first pass generates an initial prediction, which is encoded into prompt embeddings and fused with text features; the second pass refines the prediction to yield the final segmentation mask.

Training proceeds in two stages. In the first stage (\emph{positive-only pretraining}), each training instance includes two positive samples and no negative samples. The model is trained with the Muon optimizer~\cite{jordan2024muon} using a learning rate of $1 \times 10^{-4}$, weight decay of $1 \times 10^{-5}$, and 30 epochs, which takes approximately 24 hours. In the second stage (\emph{negative fine-tuning}), each training instance contains one positive and one negative sample. The text encoder is frozen, and the remaining parameters are fine-tuned with Muon using a learning rate of $1 \times 10^{-5}$, weight decay of $1 \times 10^{-6}$, and 5 epochs, which takes about 4 hours.

All experiments are conducted on 8 NVIDIA B200 GPUs with a cumulative batch size of $16 \times 8$, supported by 96 CPU cores and 1024 GB of system memory. Training uses mixed precision (bfloat16) and ZeRO-2 optimizer sharding from DeepSpeed~\cite{rasley2020deepspeed}. The learning rate follows a cosine schedule with a warmup ratio of 0.03.

\section{Results}
\label{section4}

We evaluate two variants of our model to study scalability: \textbf{ESICA} and \textbf{ESICA-Lite}. The two variants differ only in their text encoders. ESICA employs the full ClinicalBERT~\cite{wang2023optimized,liu2025generalist} for processing textual prompts, while ESICA-Lite adopts the more compact TinyClinicalBERT~\cite{rohanian2023lightweight}. All other architectural components and training hyperparameters remain identical.

\subsection{Quantitative Results}
Table~\ref{tab:overall_result} reports the quantitative performance on the CVPR-BiomedSegFM validation set. Overall, ESICA achieves the strongest results while maintaining a favorable balance between model capacity and computational cost. Although SAT attains relatively strong instance-level performance, it incurs an extremely high computational burden (4.62~T FLOPs) and the largest parameter count among all baselines, which limits its practicality for real-world applications. CAT is considerably more efficient than SAT (831.82~G FLOPs and 207.29~M parameters) but is consistently outperformed by our method across all evaluation metrics. In contrast, ESICA delivers the highest DSC (0.6588), NSD (0.6919), F1 (0.2162), and DSC-TP (0.5129) while requiring only 560.04~G FLOPs, indicating more effective utilization of computational resources compared with both CAT and SAT. Moreover, ESICA-Lite, with substantially fewer parameters (130.77~M) and lower FLOPs (540.61~G), continues to outperform CAT and achieves competitive performance relative to SAT, highlighting the efficiency and scalability of our architecture. As a lightweight baseline, Text3DSAM exhibits the smallest model size (59.30~M) and minimal computational cost (24.55~G FLOPs); however, its semantic accuracy (DSC 0.6091, NSD 0.6161) and instance-level performance (F1 0.1138, DSC-TP 0.2434) are markedly inferior. This observation indicates that reducing model size and computation alone is insufficient to ensure effective cross-modal text–volume alignment. Collectively, these results demonstrate that the proposed ESICA family achieves a more favorable trade-off among model size, computational complexity, and segmentation quality.

\begin{table}[!ht]
  \centering
  \caption{Overall performance on the CVPR-BiomedSegFM validation set. Results are reported for semantic segmentation (DSC, NSD) and instance segmentation (F1, DSC-TP), along with model size in millions of parameters and computational cost in FLOPs. Best results are in bold, and second-best are underlined.}
  \label{tab:overall_result}
  \resizebox{\columnwidth}{!}{%
    \begin{tabular}{lcc cccc}
      \toprule
      \multirow{2}{*}{Model}       & \multirow{2}{*}{Params (M)}  & \multirow{2}{*}{FLOPs}  & \multicolumn{2}{c}{Semantic} & \multicolumn{2}{c}{Instance} \\ \cmidrule{4-7}
      & & & DSC       & NSD       & F1        & DSC-TP \\ \midrule
      CAT         & 207.29    & 831.82 G  & 0.6427    & 0.6801    & 0.1878    & 0.3277 \\
      SAT         & 219.87    & 4.62 T  & 0.6335    & 0.6788    & \underline{0.2146}    & 0.3926 \\
      Text3DSAM   & \textbf{59.30}   &  \textbf{24.55 G}  & 0.6091    & 0.6161    & 0.1138    & 0.2434  \\ \midrule
      ESICA-Lite  & \underline{130.77}  &  \underline{540.61 G}  & \underline{0.6492}      & \underline{0.6818}      & 0.1898    & \underline{0.5033} \\
      ESICA       & 252.07  &  560.04 G  & \textbf{0.6588}      & \textbf{0.6919}      & \textbf{0.2162}      & \textbf{0.5129} \\
      \bottomrule
    \end{tabular}%
  }
\end{table}

Detailed modality-wise comparisons are presented in Table~\ref{tab:modality_result}. 
On CT, CAT achieves the highest semantic segmentation performance in terms of DSC and NSD, while ESICA and ESICA-Lite yield higher DSC-TP, indicating improved instance-level segmentation. SAT also performs competitively on instance metrics, whereas Text3DSAM lags behind in both boundary quality and instance localization. On MRI, ESICA consistently outperforms all baselines across semantic and instance metrics, demonstrating strong robustness for complex anatomical structures, while Text3DSAM shows notably lower instance-level performance.  For Microscopy, SAT achieves the highest F1 score; however, both ESICA variants substantially improve DSC-TP, suggesting more accurate delineation of fine-grained cellular instances, whereas Text3DSAM exhibits inconsistent instance-level behavior. 
On PET, SAT obtains the best F1 and DSC-TP, with ESICA remaining competitive and clearly outperforming CAT and Text3DSAM.  On Ultrasound, ESICA-Lite slightly surpasses ESICA in DSC and NSD, and both variants consistently outperform CAT, SAT, and Text3DSAM. Overall, ESICA demonstrates consistent and robust improvements across imaging modalities, with particularly strong gains on MRI and Microscopy, where accurate text–volume alignment and precise instance delineation are most challenging.

\begin{table}[!ht]
  \centering
  \caption{Quantitative results on the CVPR-BiomedSegFM validation set, broken down by imaging modality. Metrics are reported for semantic segmentation (DSC, NSD) and instance segmentation (F1, DSC-TP). “N/A” indicates metrics not applicable for a given modality. Best results are in bold, and second-best are underlined.}
  \label{tab:modality_result}
  \resizebox{\columnwidth}{!}{%
    \begin{tabular}{cl cccc}
      \toprule
      \multirow{2}{*}{Modality}  & \multirow{2}{*}{Model} & \multicolumn{2}{c}{Semantic} & \multicolumn{2}{c}{Instance} \\ \cmidrule{3-6}
      & & DSC       & NSD       & F1        & DSC-TP \\ \midrule
      \multirow{4}{*}{CT}          & CAT          & \textbf{0.7211}   & \textbf{0.7227}   & \textbf{0.2993}   & 0.3717 \\
      & SAT          & 0.678     & \underline{0.6726}    & \underline{0.2517}    & 0.3954 \\
      & Text3DSAM    & 0.6707    & 0.6095    & 0.1159    & 0.0523    \\
      & ESICA-Lite   & 0.6605    & 0.6428    & 0.2021    & \underline{0.4133} \\
      & ESICA        & \underline{0.6796}    & 0.6609    & 0.2256    & \textbf{0.4179} \\ \midrule
      \multirow{4}{*}{MRI}         & CAT          & 0.5415   & 0.6193   & 0.1375   & 0.2813 \\
      & SAT          & 0.561     & 0.6669    & 0.1228    & 0.2728 \\
      & Text3DSAM    & 0.5214    & 0.5920    & 0.1206    & 0.2519 \\
      & ESICA-Lite   & \underline{0.6050}    & \underline{0.6871}    & \underline{0.2145}    & \underline{0.4971} \\
      & ESICA        & \textbf{0.6082}    & \textbf{0.6921}    & \textbf{0.2444}    & \textbf{0.5047} \\ \midrule
      \multirow{4}{*}{Microscopy}  & CAT          & N/A       & N/A       & 0.0313   & 0.3628 \\
      & SAT          & N/A      & N/A      & \textbf{0.2006}    & 0.4243 \\
      & Text3DSAM    & N/A       & N/A       & 0.0386    & 0.6918    \\
      & ESICA-Lite   & N/A      & N/A       & 0.1157    & \underline{0.6307} \\
      & ESICA        & N/A      & N/A       & \underline{0.1452}    & \textbf{0.6554} \\ \midrule
      \multirow{4}{*}{PET}         & CAT          & N/A       & N/A       & 0.1098   & 0.2779 \\
      & SAT          & N/A      & N/A      & \textbf{0.42}      & \textbf{0.7863} \\
      & Text3DSAM    & N/A       & N/A       & 0.0886    & 0.4484    \\
      & ESICA-Lite   & N/A      & N/A       & 0.1381    & 0.7384 \\
      & ESICA        & N/A      & N/A       & \underline{0.1593}    & \underline{0.7578} \\ \midrule
      \multirow{4}{*}{Ultrasound}  & CAT          & 0.8594   & 0.836   & N/A       & N/A    \\
      & SAT          & 0.8558    & 0.7924    & N/A       & N/A    \\
      & Text3DSAM    & 0.8337    & 0.8117    & N/A       & N/A    \\
      & ESICA-Lite   & \textbf{0.8738}    & \textbf{0.8699}    & N/A      & N/A    \\
      & ESICA        & \underline{0.8720}    & \underline{0.8672}    & N/A      & N/A    \\ \midrule
      \multirow{4}{*}{Average}     & CAT          & 0.6427   & 0.6801   & 0.1878    & 0.3277  \\
      & SAT          & 0.6335    & 0.6788    & \underline{0.2146}      & 0.3926    \\
      & Text3DSAM    & 0.6091    & 0.6161    & 0.1138    & 0.2434  \\
      & ESICA-Lite   & \underline{0.6492}    & \underline{0.6818}    & 0.1898      & \underline{0.5033}    \\
      & ESICA        & \textbf{0.6588}   & \textbf{0.6919}    & \textbf{0.2162}       & \textbf{0.5129}    \\
      \bottomrule
    \end{tabular}%
  }
\end{table}

\subsection{Qualitative Results}
\begin{figure*}[!hbt]
  \centering
  \includegraphics[width=0.85\textwidth]{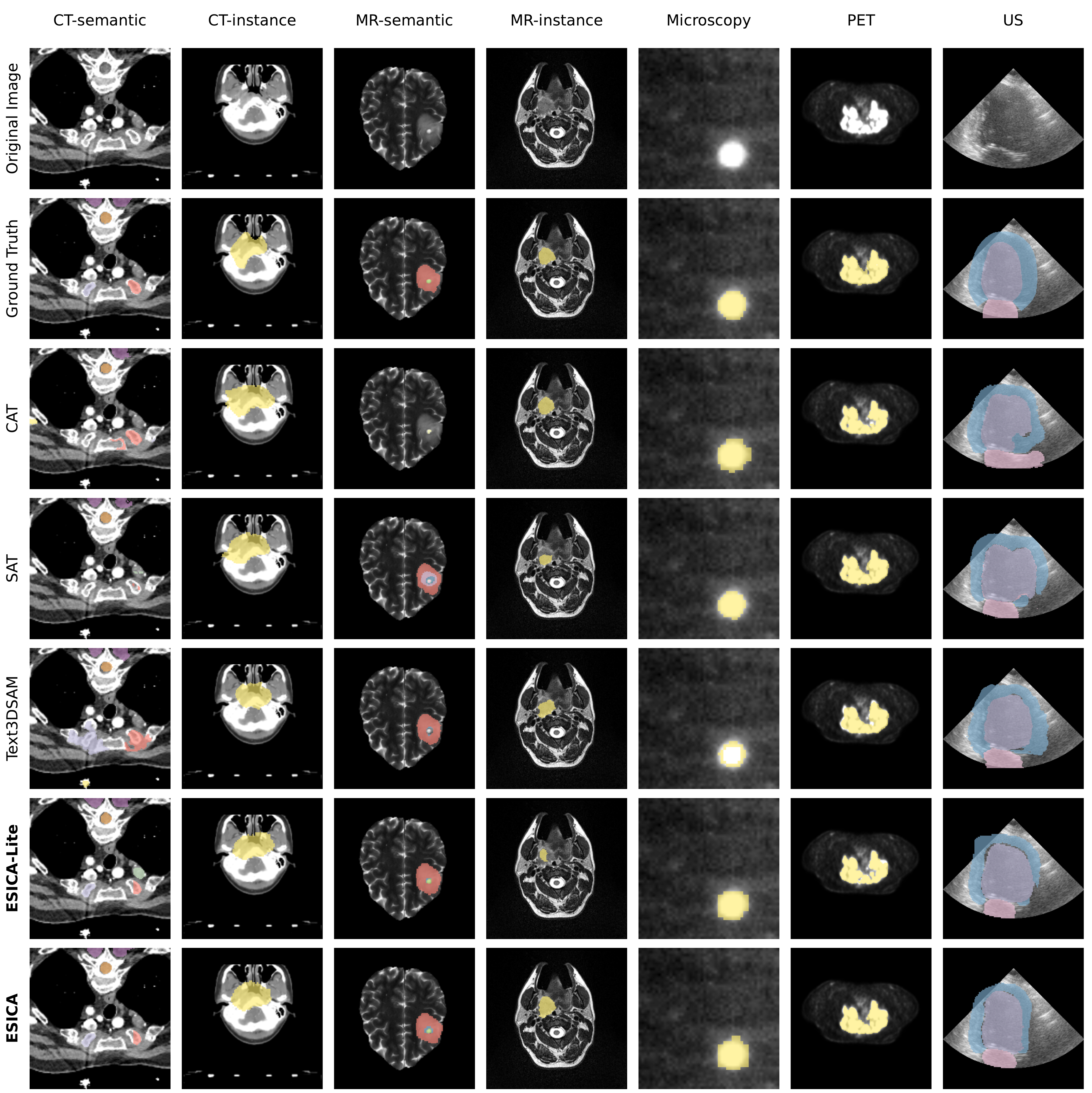}
  \caption{Qualitative segmentation results across modalities on the CVPR-BiomedSegFM validation set.}
  \label{fig:qual_result}
\end{figure*}

Figure \ref{fig:qual_result} illustrates representative segmentation results across modalities. CAT and SAT frequently struggle to capture fine anatomical boundaries, often leading to under-segmentation or missing small structures. By contrast, ESICA produces sharper boundaries and more accurate delineations, particularly in MRI and Microscopy. ESICA-Lite, despite its reduced size, achieves visual quality comparable to ESICA and demonstrates robust performance.

In CT and MRI semantic segmentation, both ESICA variants provide clearer and more precise boundaries, with MRI showing the most notable improvements in fine anatomical detail. For CT instance segmentation, performance is comparable across models, whereas in MRI instance segmentation, ESICA and CAT outperform ESICA-Lite and SAT. In Microscopy, ESICA-Lite performs competitively despite using a smaller text encoder, producing only minor false positives at edges while maintaining true positive regions comparable to CAT and ESICA. In Ultrasound, ESICA achieves the best results, with ESICA-Lite slightly outperforming CAT and SAT.

\section{Ablation Study}
\label{section5}

We conduct ablation experiments to evaluate the impact of training strategies, architectural components, and refinement iterations in ESICA.

\subsection{Impact of Training Strategies and Architecture}

Table~\ref{tab:ablation_module} reports the results of selectively removing key training strategies and architectural modules. Both negative fine-tuning and positive-only pretraining are essential for achieving strong performance. Excluding negative fine-tuning reduces DSC from 0.6588 to 0.5442 and NSD from 0.6919 to 0.5472, suggesting that hard negative signals help sharpen decision boundaries. Removing positive-only pretraining causes a severe performance collapse, with DSC dropping to 0.2123 and F1 to 0.0276, confirming that this stage is indispensable for stable cross-modal alignment.

The similarity alignment mask also plays a significant role: omitting it decreases DSC from 0.6588 to 0.5854 and DSC-TP from 0.5129 to 0.4685, reflecting weaker localization of fine anatomical structures. Finally, replacing DCFormerV2 with the original DCFormer reduces DSC to 0.6396 and DSC-TP to 0.4797, demonstrating the benefit of the additional \(3 \times 3 \times 3\) convolution branch and enhanced transformer design in DCFormerV2.

\begin{table}[!ht]
  \centering
  \caption{Ablations on training strategies and architectural components of ESICA on the CVPR-BiomedSegFM validation set.}
  \label{tab:ablation_module}
  \resizebox{\columnwidth}{!}{%
    \begin{tabular}{l cccc}
      \toprule
      \multirow{2}{*}{Model}       & \multicolumn{2}{c}{Semantic} & \multicolumn{2}{c}{Instance} \\ \cmidrule{2-5}
      & DSC       & NSD       & F1        & DSC-TP \\ \midrule
      ESICA       & \textbf{0.6588}      & \textbf{0.6919}      & \textbf{0.2162}      & \textbf{0.5129} \\ \midrule
      \textit{w/o negative fine-tuning} & 0.5442 & 0.5472 & 0.1429 & 0.5280 \\
      \textit{w/o positive-only pretraining} & 0.2123 & 0.2305 & 0.0276 & 0.0943 \\ \midrule
      \textit{w/o similarity alignment mask} & 0.5854 & 0.6011 & 0.1480 & 0.4685 \\
      \textit{w/o DCFormerV2} & 0.6396 & 0.6817 & 0.2012 & 0.4797 \\
      \bottomrule
    \end{tabular}%
  }
\end{table}

\subsection{Impact of Refinement Iterations}

Table~\ref{tab:ablation_iteration} evaluates the effect of iterative refinement. A single-pass prediction performs poorly (DSC 0.4185, F1 0.0334), indicating the difficulty of accurate segmentation without refinement. Introducing a second pass substantially improves performance, increasing DSC to 0.6588, NSD to 0.6919, and F1 to 0.2162; we therefore adopt this setting as the default configuration. Adding a third pass does not yield further gains and slightly degrades DSC (0.6328) and NSD (0.6643), likely due to error accumulation across iterations. These results validate the use of two refinement iterations as an effective balance between accuracy and computational efficiency.

\begin{table}[!ht]
  \centering
  \tiny
  \caption{Ablations on refinement iterations of ESICA on the CVPR-BiomedSegFM validation set.}
  \label{tab:ablation_iteration}
  \resizebox{\columnwidth}{!}{%
    \begin{tabular}{l cccc}
      \toprule
      \multirow{2}{*}{Model}       & \multicolumn{2}{c}{Semantic} & \multicolumn{2}{c}{Instance} \\ \cmidrule{2-5}
      & DSC       & NSD       & F1        & DSC-TP \\ \midrule
      ESICA (1-pass) & 0.4185 & 0.3728 & 0.0334 & 0.2434 \\
      ESICA (2-pass)       & \textbf{0.6588}      & \textbf{0.6919}      & \textbf{0.2162}      & \textbf{0.5129} \\
      ESICA (3 pass) & 0.6328 & 0.6643 & 0.2104 & 0.4890 \\
      \bottomrule
    \end{tabular}%
  }
\end{table}

\section{Discussion}
\label{section6}

This study introduces ESICA, a text-guided framework for 3D medical image segmentation that emphasizes semantic alignment, computational efficiency, and scalability. Across a large and diverse benchmark, ESICA consistently achieves strong segmentation performance while using fewer computational resources than existing text-guided methods. These results indicate that accurate text-guided volumetric segmentation can be achieved through principled alignment and efficient design, rather than increased model scale alone.

A key factor underlying ESICA’s performance is the similarity-based mask prediction strategy, which directly links dense image features to textual representations. This formulation enables the model to localize regions that correspond closely to the semantic intent of the text prompt, without relying on additional supervision or complex loss terms. The resulting improvements are most evident in modalities such as MRI and Microscopy, where anatomical structures are small, irregular, or weakly contrasted. These findings suggest that explicit image–text similarity modeling is an effective and general mechanism for text-guided 3D segmentation.

ESICA also achieves a favorable balance between accuracy and efficiency. Compared with prior approaches, it delivers higher or comparable segmentation quality with substantially lower computational cost. This efficiency is enabled by lightweight architectural components, including the enhanced DCFormerV2 encoder, an efficient decoder with adapter modules, and streamlined cross-modal attention. Importantly, the strong performance of the ESICA-Lite variant shows that high-quality text-guided segmentation remains feasible even under constrained model capacity, which is critical for practical clinical deployment.

The iterative refinement strategy further contributes to segmentation quality. A single-pass prediction is often insufficient for complex 3D structures, particularly when guided by free-form text. Introducing a second refinement pass allows the model to correct coarse boundaries and ambiguous regions using feedback from the initial prediction, leading to substantial performance gains. Additional refinement iterations do not provide further benefit and may introduce instability, indicating that limited refinement strikes the best balance between accuracy and efficiency.

The training strategy plays an equally important role. Positive-only pretraining establishes stable image–text alignment by avoiding early exposure to ambiguous negative samples, while subsequent balanced fine-tuning improves discrimination and generalization. The sharp performance degradation observed when either stage is removed highlights the importance of curriculum design in large-scale text-guided segmentation.

Despite these strengths, several limitations remain. Instance-level detection performance, as measured by F1 score, lags behind voxel-level segmentation quality, suggesting that identifying object presence remains more challenging than delineating object boundaries. This limitation likely stems from limited negative sampling during fine-tuning and could be addressed through more targeted negative mining or instance-aware objectives. In addition, while the current cross-modal interaction design is effective, alternative fusion strategies may further improve robustness to complex or compositional language prompts.

\section{Conclusion}
This paper presents ESICA, an efficient and scalable framework for text-guided 3D medical image segmentation that directly aligns volumetric image features with natural language descriptions. By combining similarity-based mask prediction, lightweight architectural design, iterative refinement, and a structured training strategy, ESICA achieves strong segmentation performance across diverse imaging modalities while maintaining a favorable balance between accuracy and computational efficiency. Extensive experiments demonstrate that ESICA consistently outperforms existing text-guided approaches and remains effective even in compact configurations, highlighting its potential for practical deployment. Overall, this work advances text-guided segmentation toward more flexible, efficient, and clinically applicable vision–language systems.

\bibliographystyle{IEEEtran}
\bibliography{ref}

\end{document}